# Robotic Assistant Agent for Student and Machine Co-Learning on AI-FML Practice with AIoT Application

Chang-Shing Lee, Mei-Hui Wang
Zong-Han Ciou, Rin-Pin Chang
Chun-Hao Tsai
National University of Tainan
Tainan, Taiwan
leecs@mail.nutn.edu.tw

Shen-Chien Chen
Zsystem Technology Co.
Kaohsiung, Taiwan
xol700@gmail.com

Tzong-Xiang Huang
Eri Sato-Shimokawara
Toru Yamaguchi
Tokyo Metropolitan University
Tokyo, Japan
yamachan@tmu.ac.jp

*Abstract*—**In this paper, the Robotic Assistant Agent for student and machine co-learning on AI-FML practice with AIoT application is presented. The structure of AI-FML contains three parts, including fuzzy logic, neural network, and evolutionary computation. Besides, the Robotic Assistant Agent (RAA) can assist students and machines in co-learning English and AI-FML practice based on the robot Kebbi Air and *AIoT-FML learning tool*. Since Sept. 2019, we have introduced an Intelligent Speaking English Assistant (ISEA) App and AI-FML platform to English and computer science learning classes at two elementary schools in Taiwan. We use the collected English-learning data to train a predictive regression model based on students' monthly examination scores. In Jan. 2021, we further combined the developed AI-FML platform with a novel *AIoT-FML learning tool* to enhance students' interests in learning English and AI-FML with basic hands-on practice. The proposed RAA is responsible for reasoning students' learning performance and showing the results on the *AIoT-FML learning tool* after communicating with the AI-FML platform. The experimental results and the collection of students' feedback show that this kind of learning model is popular with elementary-school and high-school students, and the learning performance of elementary-school students is improved.**

## I. INTRODUCTION

Computational Intelligence (CI), including fuzzy logic, neural network, and evolutionary computation, is a sub-branch of AI. It is an important core technology of AI and plays an important role in developing successful intelligent systems, including games, multilayer perceptron, and cognitive developmental systems [1]. Fuzzy logic is suitable for computing the degree of human perception such as hot or cold. Different people have different feelings of hot and cold even at the same temperature. The neural network is one of the important models for machine learning which can compute the mathematical feature functions. Evolutionary computation is based on the observation of the animals' behavior patterns and it is one of the important machine learning models, too.

Fig. 1 shows the structure of AI-FML human and machine hands-on practice. AI-FML contains three parts, including fuzzy logic such as human knowledge and logic operation rule, a neural network such as machine learning model and deep learning architecture, and evolutionary computation such as genetic algorithm and particle swarm optimization. The learning topics of AI-FML are composed of two phases: 1) *Before Machine Learning*: *AI-FML human and machine hands-on practice* contains human knowledge construction and data collection & analysis. 2) *After Machine Learning*: *AI-FML human and machine cooperation* contains the construction of machine learning model and applications of the learned model in everyday life.

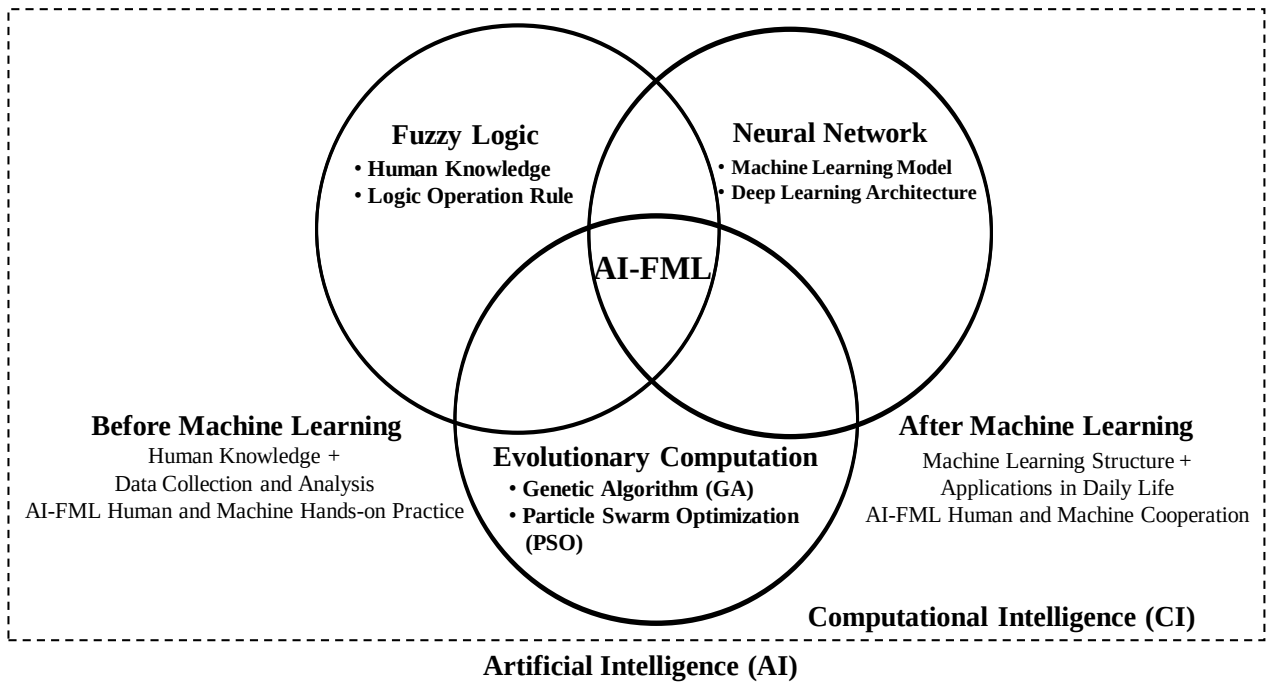

Fig. 1. Structure of AI-FML human and machine hands-on practice.

Owing to the advent of technologies, many AIoT devices are applied to many areas such as smart living, smart manufacturing, smart agriculture, smart health, and smart city [2]. There exist some research publications, for example, Zhang et al. [3] proposed an AIoT-based system for real-time monitoring of tunnel construction. Chiu et al. [4] developed a DNN based emotionally aware campus virtual assistant to help users find information in complex campus web pages. Lee et al. [5, 7, 8] deployed the AI-FML agent to AIoT devices such as the robot Kebbi Air, and it has been introduced to the learning course of 5-grade computer studies at Rende elementary school in Taiwan since Sept. 2019. In this way, the involved students can co-learn AI-FML with the robot based on the MQTT transmission protocol.

FML was first proposed by Loia and Acampora in 2004 [12] and has become IEEE 1855 and its available software tools are as follows: (1) VisualFMLTool : It can be executed on platforms containing Java Runtime Environment. The Java Software Development Kit, including JRE, compiler, and many other tools can be found here. The VisualFMLTool can download from here and then extract. Then it is only needed to click the file VisualFMLTool.bat included in the zip to execute the tool. (2) AI-FML Tool: It is developed by KWS center/OASE Lab., NUTN, Taiwan and can be executed on different platforms online. After registering for the competition, we can provide an account for the participants [8]. (3) JFML: A Spanish research group (Jose Manuel Soto Hidalgo, Giovanni Acampora, Jesus

The authors would like to thank the financial support sponsored by the Ministry of Science and Technology of Taiwan under the grants MOST 109-2622-E-024-001-CC1.

Alcala Fernandez, Jose Alonso Moral) has released a library for FML programming that is very simple to use and compliant with IEEE 1855. JFML can download from here. Additional information about the library is here [13, 14, 15].

In this paper, we propose a *Robotic Assistant Agent* (*RAA*) for student and machine co-learning English and AI-FML practice with AIoT application. The involved elementary-school students first speak English to the robot Kebbi Air through the developed *Intelligent Speaking English Assistant* (*ISEA*) app. After that, the *ISEA* provides the speaking performance for the students such as saying words of encouragement and displaying a score recognized by the *RAA*. In addition to the robot Kebbi Air, we first developed and introduced an *AIoT-FML learning tool* based on the GenioPy board [6], to the AI-FML club class in Dec. 2020 to escalate their learning interests. Through gathering the observations and measurements, we analyze their learning performance based on the deep learning architecture. In addition to elementary-school students, we cooperated with Taiwan AI Academy to export this kind of learning AI-FML model to the AI experience camp of senior-high-school students in Feb. 2021. From the collected feedback, learning AI-FMLfor future real-world application is welcome by most high school students.

The remainder of this paper is organized as follows. The structure of the robotic assistant agent for student and machine co-learning is introduced in Section II. Then, the proposed AI-FML knowledge model on AIoT application for student learning is described in Section III. Section IV shows some experimental results. Finally, we conclude this paper in Section V.

## II. Structure of Robotic Assistant Agent for Student and Machine Co-Learning

### A. *Human & Machine Cooperation, Co-Learning, and Symbiosis*

Fig. 2 shows a scenario of AI-FML human & machine cooperation, co-learning, and symbiosis described as follows [9, 10]:

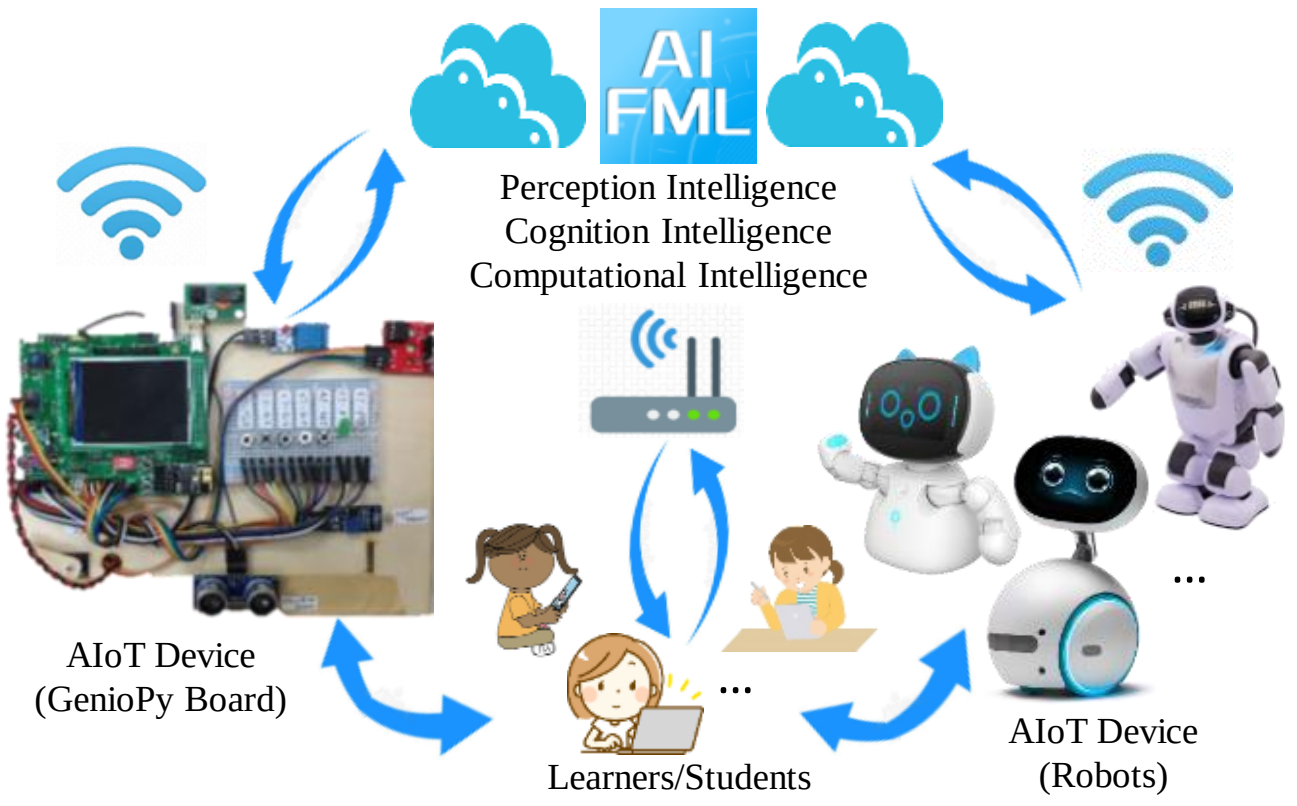

Fig. 2. A scenario of AI-FML human & machine cooperation, co-learning, and symbiosis [9].

1) Learners/students surf on the AI-FML platform to construct human knowledge and logic operation rules of applications to daily life. 2) Learners/Students interact with the AIoT devices to co-learn with the machines to collect the data of the observations. 3) AI-FML with the perception intelligence, cognition intelligence, and computational intelligence can send the inferred results to *AIoT-FML learning tools* such as GenioPy board and robots based on the input data and the constructed KB/RB. 4) Implement the approaches of machine learning to train the learning model and predict future learning performance. 5) Repeat Steps 1 to 4 to achieve the goal of AI-FML human & machine cooperation, co-learning, and symbiosis. Based on the concepts of the AI-FML human & machine cooperation, co-learning, and symbiosis, we will hold an FML-based machine learning competition for human and smart machine co-learning on real-world applications @ FUZZ-IEEE 2021 in July 2021 [9, 10].

### B. *Robotic Assistant Agent for Student English Learning Structure*

Fig. 3 shows the structure of the proposed *Robotic Assistant Agent* (*RAA*) for student English learning. In this structure, students can learn the AI-FML practice (machine language) and English (human language) to achieve the goal of human and machine co-learning. We give some descriptions as follows: 1) In the teaching fields, the left-side students first sign in the *Intelligent Speaking English Assistant* (*ISEA*) app, choose a suitable sentence from the teaching materials to practice, and speak the selected sentence to the robot Kebbi Air. The right-hand students observe the information shown on the *AIoT-FML learning tool*, such as LCD of the GenioPy board. 2) The collected data, including how many times they have practiced speaking, what sentences they speak, all AI-recognized scores, and so on, are stored in the *ISEA* server. 3) Students log in to the AI-FML platform to build up their own KB/RB for the *ISEA* application. After that, they input values for the constructed KB/RB to infer the results. 4) the *AIoT-FML learning tool*, such as the GenioPy board, shows the inferred result together with saying something through communication with the AI-FML server while Kebbi Air also communicates with the MQTT server to take some actions to match with the inferred result. 5) Students use their data of speaking English to train their English learning model and learn the machine learning language, AI-FML.

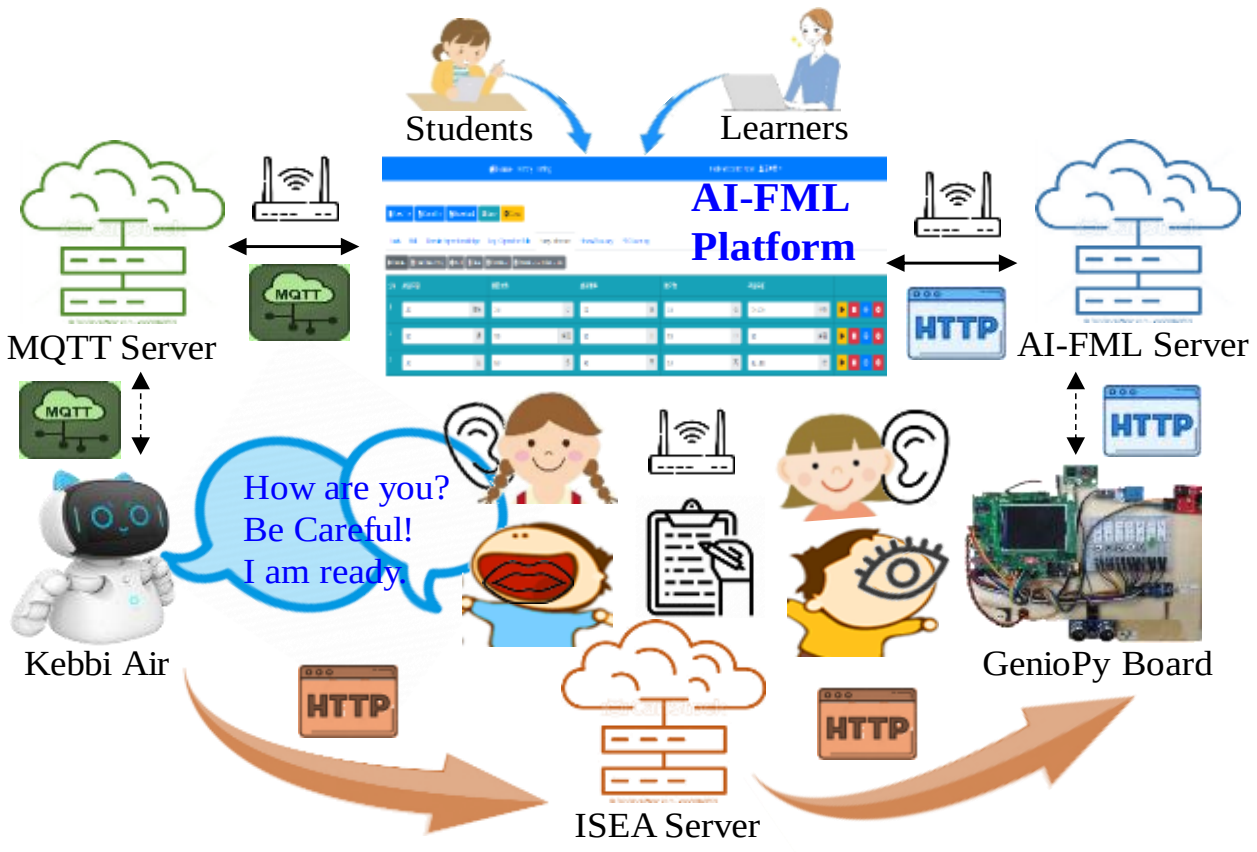

Fig. 3. Robotic assistant agent for student English learning structure.

### C. *Robotic Assistant Agent on AIoT Application*

In this paper, we cooperated with Mediawave Intelligent Communication Limited. and Zsystem Technology Co. to construct the *AIoT-FML learning tool* and make the GeniopPy board be with basic ability in intelligence by installing the *RAA* and embedding FML function in it. In this way, the users can apply AI-FML with the features of human intelligence and artificial intelligence to real-world applications. Fig. 4(a) is the screenshot of the *AIoT-FML learning tool* based on the GenioPy board [6]. It has the following features: 1) ARM926EJ-S CPU with both 16K-byte unified I/D-cache, 64KB TCM, embedded JTAG ICE, and working frequency up to 513MHz. 2) Built-in Picture Process Unit (PPU) for Image processing and Game Function, Baseline JPEG/Motion JPEG Video Encoder/Decoder. 3) Built-in Sound Process Unit (SPU), Dynamic Volume Compressor, MP3/WAV Encoder/Decoder. 4) Edge Computing Voice Recognition (EdCVR). 5) Edge Computing Face Recognition/Object Recognition (EdCFR/EdCOR). 6) *AI-FML: Human Intelligence & Artificial Intelligence for Real-World Applications*.

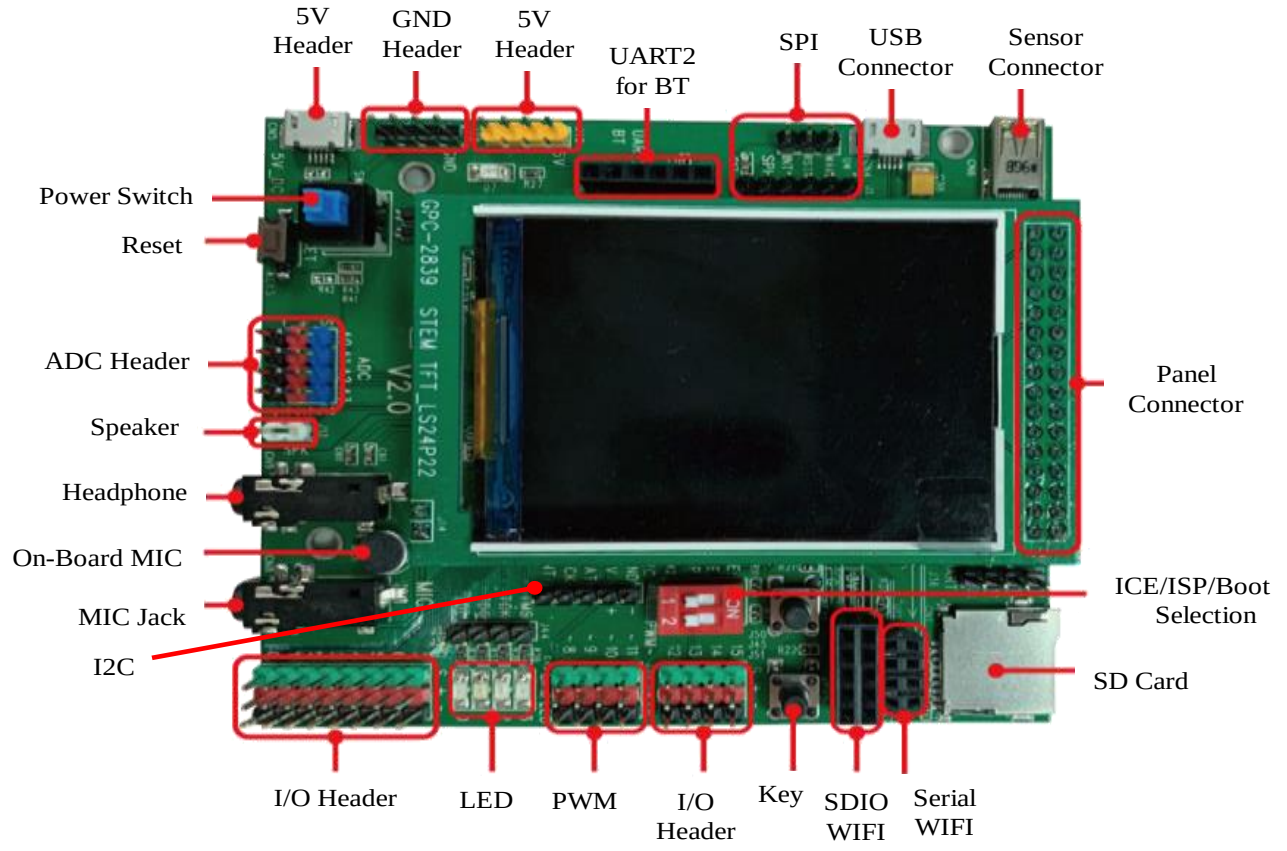

(a)

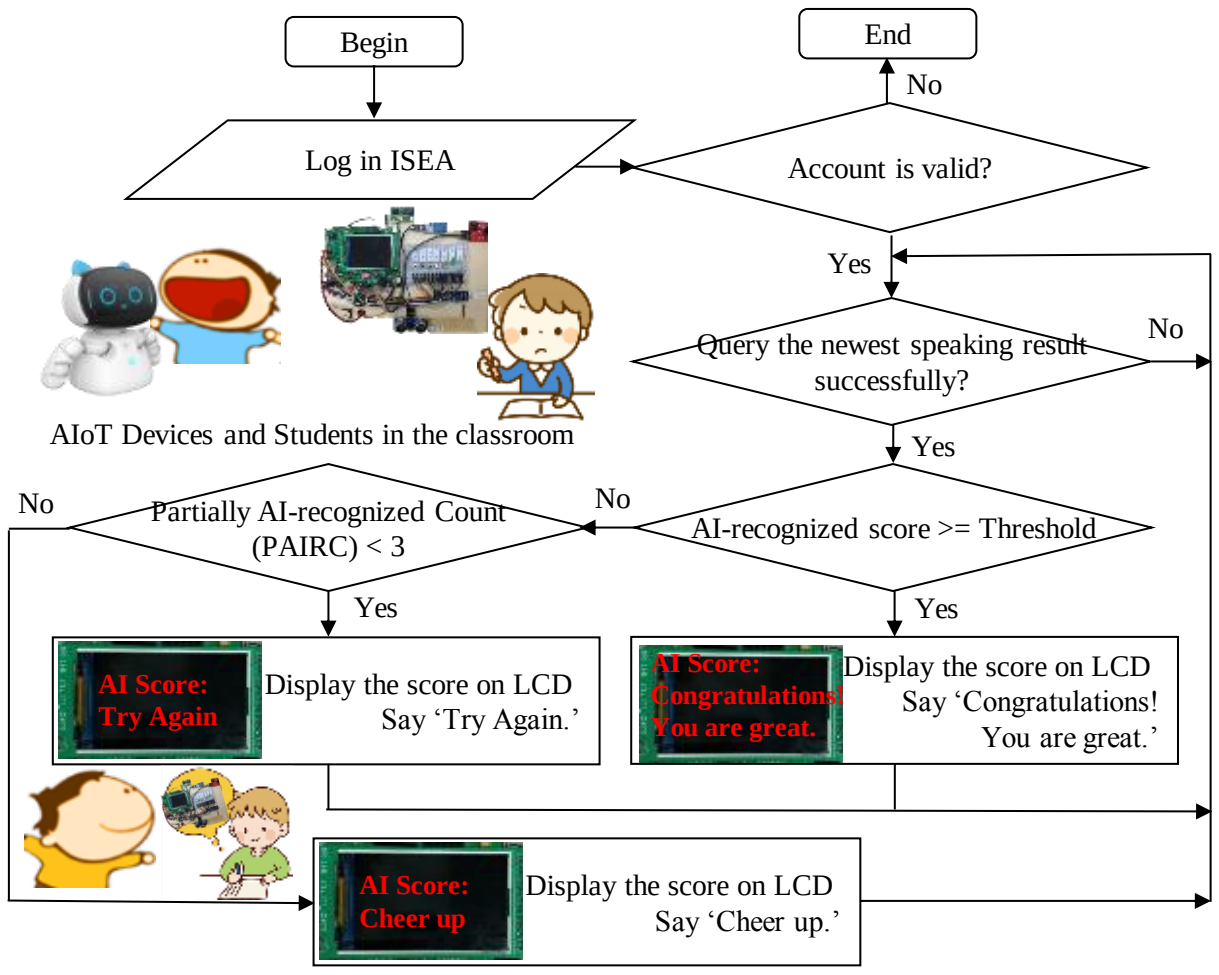

(b)

Fig. 4. (a) Screenshot of the *AIoT-FML learning tool* based on GenioPy board and (b) the *RAA* flowchart for student English learning.

Fig. 4(b) shows the *RAA* flowchart for elementary-school or high-school student English learning. There are many students in the classroom; some are speaking English to the robot Kebbi Air, others are observing the information on the *AIoT-FML learning tool* based on the GenioPy board, and still, others are writing some notes such as the acquired *RAA*-recognized score. The *ISEA* server receives the students' speeches and recognizes them to decide a corresponding fuzzy score. If this fuzzy score is greater than the predefined threshold, where we set 0.5 in this paper, the *ISEA* server gives a correctly recognized score and the *RAA* says "*Congratulations! You are great.*" On the contrary, the *ISEA* server gives a partially *RAA*-recognized score. At this time, if partially *RAA*-recognized count (*PAIRC*) is less than 3, the *RAA* says "*Try Again*." However, if *PAIRC* is equal to 3 or greater than 3, then the *RAA* says "*Cheer Up*."

## III. AI-FML Knowledge Model for Student Learning on AIoT Application

### *A. AI-FML Knowledge Model for Student Learning*

In this section, we describe the AI-FML knowledge model for student learning on AIoT applications and its testing model between Taiwan and Japan. Fig. 5 shows the structure of the AI-FML knowledge model, including AI-FML learning and practice platform, embedded AI-FML robots, and *AIoT-FML learning tool* for student learning and practice in the classroom, especially for elementary-school and high-school students. There are three types of AI-FML knowledge tools for student learning, including the *AI-FML learning and practice platform*, *embedded AI-FML robots* like Kebbi Air for real-world applications, and *AIoT-FML learning tools* such as GenioPy board, for high school or undergraduate students learning and experiments. Students can construct the knowledge base (KB) and rule base (RB) for their fuzzy inference system, then apply machine learning models such as genetic algorithm (GA), particle swarm optimization (PSO), or neural network for AI-FML system optimization. The involved students can co-learn English with the robots such as Zenbo and Zenbo Junior by speaking or listening.

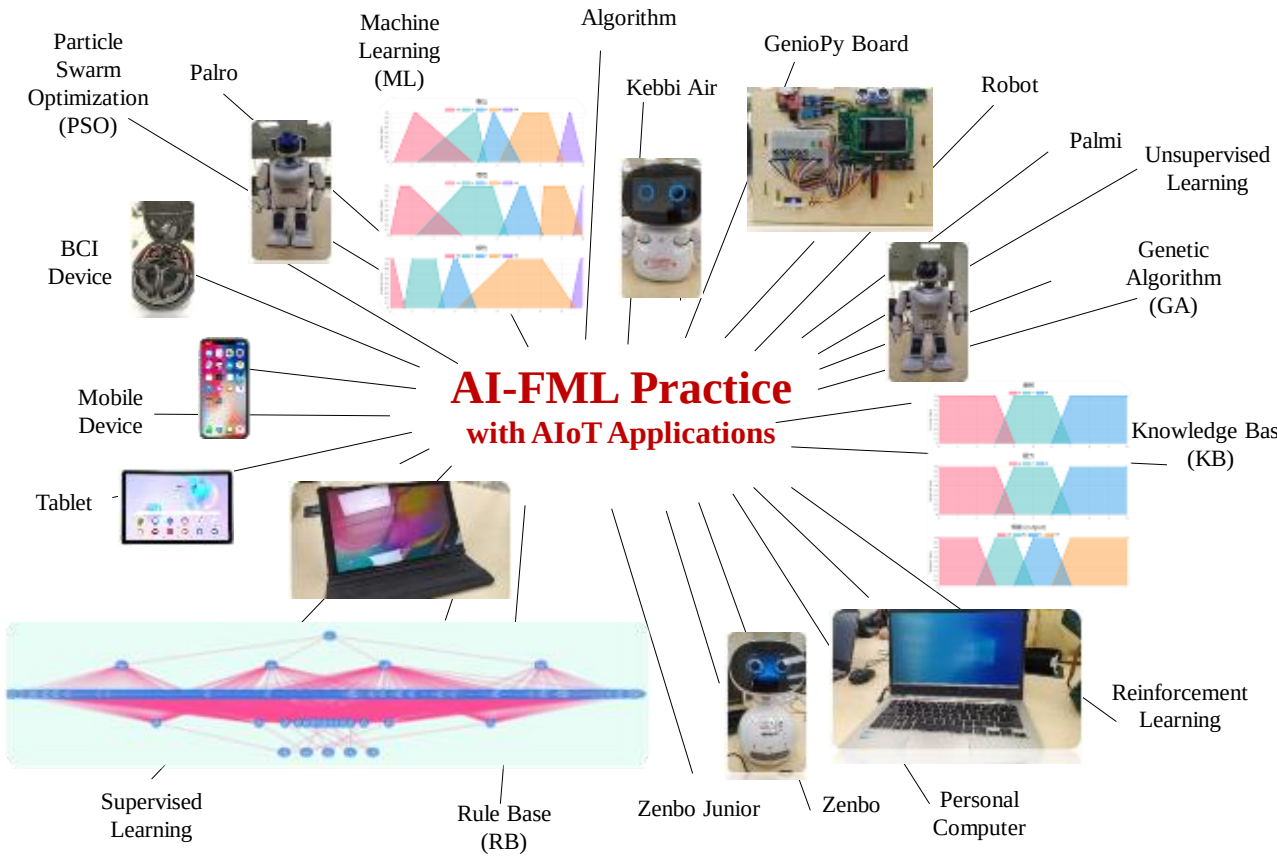

Fig. 5. Structure of AI-FML knowledge model for student learning.

### *B. AI-FML for English Speaking with AIoT Applications and Testing between Japan and Taiwan*

Fig. 6 shows the testing scenario for human English and AI-FML learning with AIoT applications described as follows: 1) The involved users first log in to Google Meet in Taiwan and Japan. A student in Hino campus, Tokyo Metropolitan University (TMU), Japan speaks English with the robot Kebbi Air. 2) The developed *RAA* online evaluates how well he speaks, then the *AIoT-FML learning tool* in South Campus, Taiwan AI

Academy, Taiwan receives an evaluation of the results and shows his performance of human and robot co-learning. 3) The student in Japan gives a command to the robot Palmi/Palro in Taiwan to do what he wants the robot to do. Based on this learning model, the high-school or elementary-school students in the teaching field can learn and practice how to collect and record human-learning data in preparation for machine learning. For more details in the testing, please connect to https://youtu.be/hcZ-Fd7WC1s. Under this scenario, people in the world can co-learn AI-FML with the learning platform and AIoT-FML learning tool together.

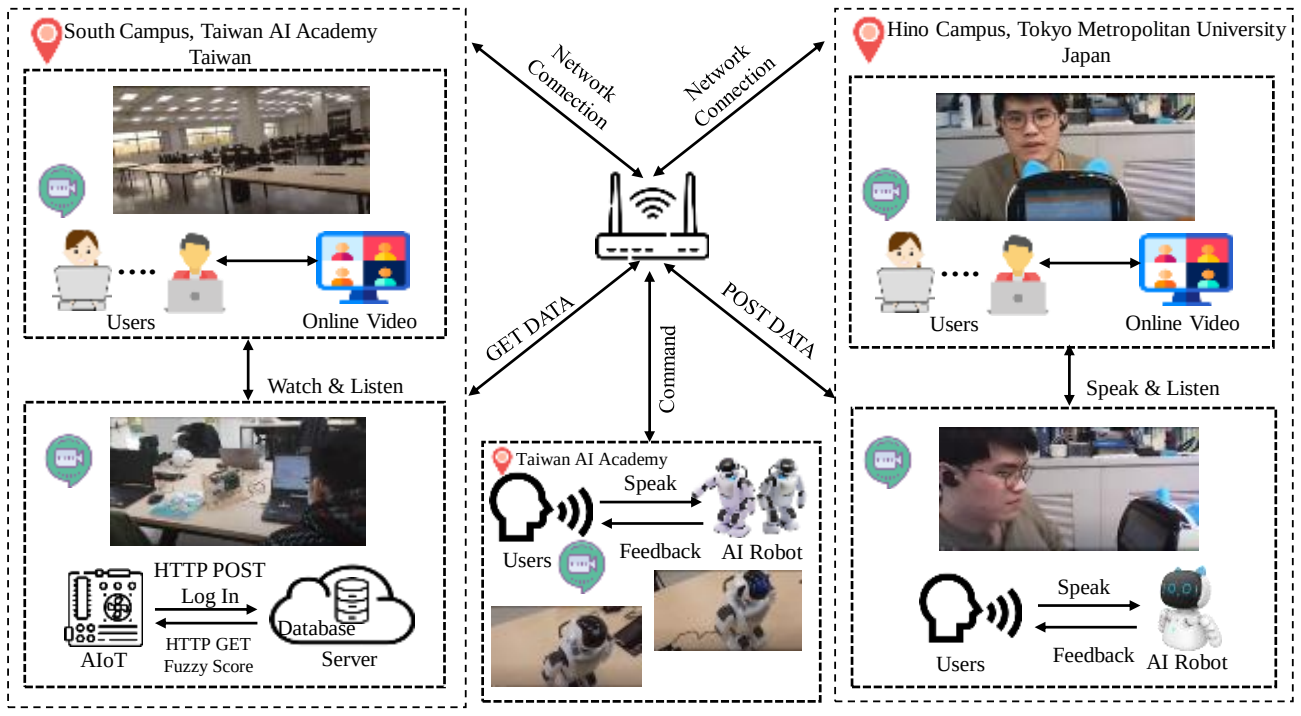

Fig. 6. The testing scenario for human English and AI-FML learning with *AIoT-FML learning tool* between Japan and Taiwan.

## C. *RAA Embedded AI-FML Robot with AIoT-FML learning tool for Teacher Teaching*

The teaching contents of the knowledge model of AI-FML for elementary-school or high-school students in Taiwan include: 1) Intelligence vs. Artificial Intelligence, 2) Weak AI vs. Human-Level AI, 3) Brain-Computer Interface (BCI) & Real-World Applications, 4) AI Development Videos, 5) Human & Machine Co-Learning Practices, and 6) AI-FML Activity @ IEEE. Fig. 7 shows the *RAA* embedded AI-FML robot with *AIoT-FML learning tool* for teacher teaching, as well as students and machine co-learning in traditional computer classrooms and AI classrooms. There are two kinds of classrooms, including a computer classroom and an AI classroom. When students interact with the AI-FML robots or *AIoT-FML learning tool*, they take an AI-FML course in the AI classroom. If they learn how to operate the AI-FML platform to construct KB and RB, analyze the collected data, or execute the tools of machine learning, they take the course in a computer classroom with a more stable Internet.

Fig. 8 shows AI-FML learning and practice course design for high school or elementary school students. We describe as follows: 1) Teachers first explain the learning goals of the class. 2) Teachers and students prepare and set up the *AI-FML robots*, *AIoT-FML learning tool*, or the other equipment in the classroom. 3) Each student speaks five to ten English sentences to *AI-FML robots*. 4) The other students can observe, collect, and record the speaking score data based on the *AIoT-FML learning tool* from GenioPy Board for the AI-FML machine learning mechanism. 5) Students organize and analyze the collected data. 6) The particle swarm optimization (PSO) algorithm is utilized for the AI-FML machine learning model in this paper. The PSO-based AI-FML model learns each student's learning performance and sends the evaluation result to the student and his/her teacher.

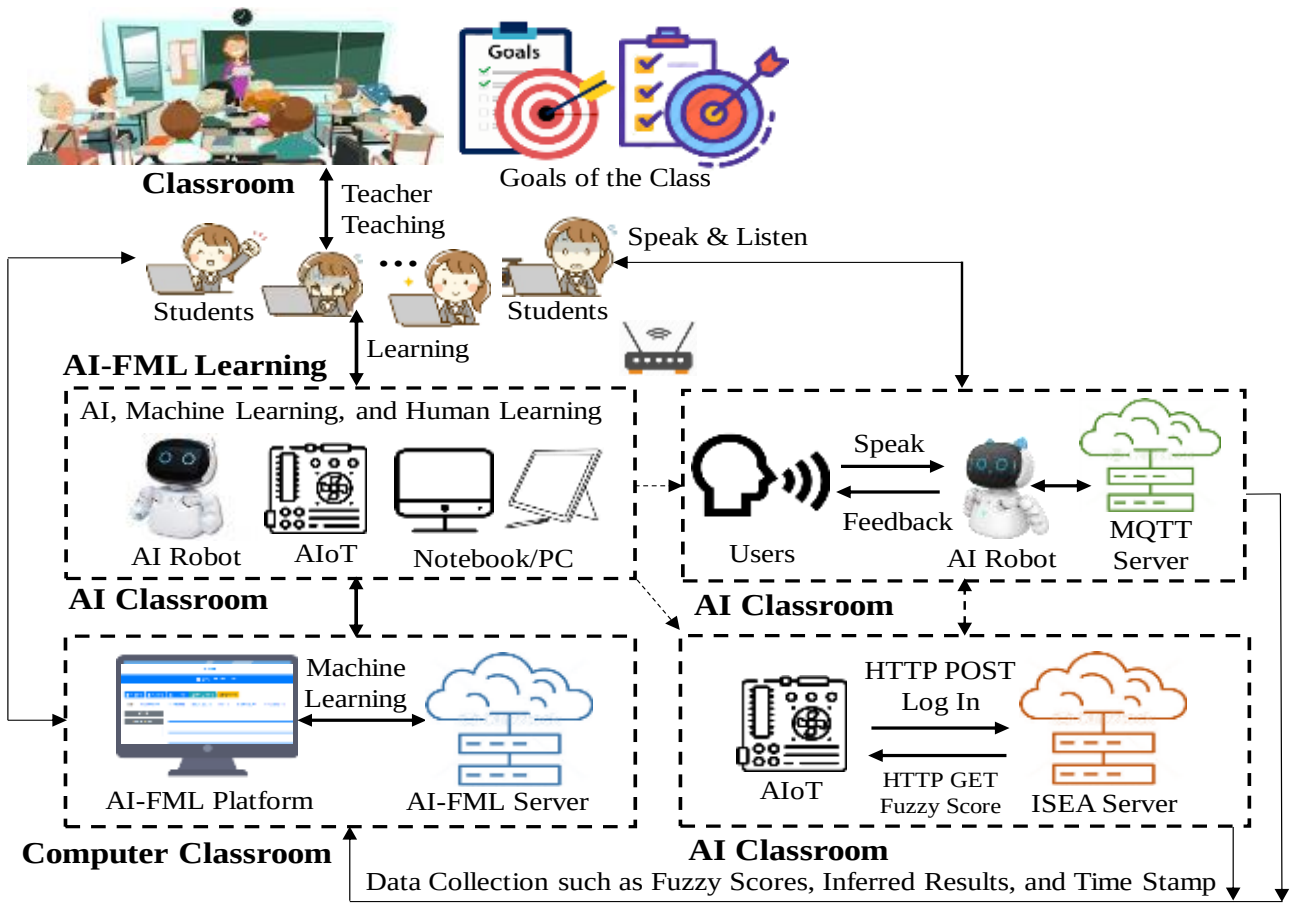

Fig. 7. *RAA* for teacher teaching, student learning, and machine co-learning based on the *AIoT-FML learning tool* in a traditional classroom.

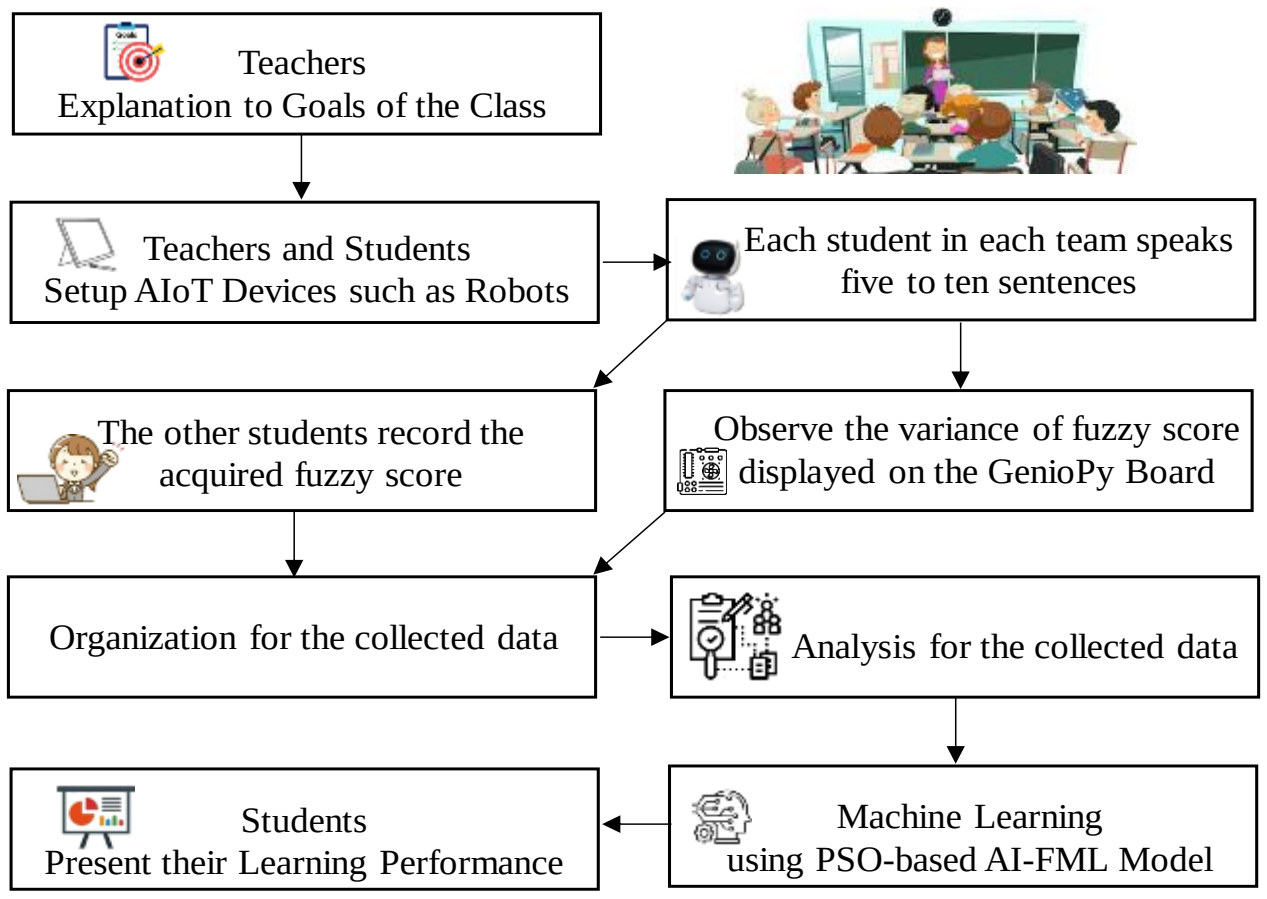

Fig. 8. The AI-FML learning and practice course design for high school or elementary school students.

# IV. Experimental Results

## A. *ISEA Data Analysis at English Co-Learning Class*

To evaluate the effectiveness and performance of the *RAA* for student and machine co-learning on English speaking with AIoT applications, we invite elementary-school students, junior-high-school students, and senior-high-school students to join the learning course and winter camp in Taiwan.

TABLE I. Valid English Speaking Data.

| Data Set | Period of Semester | Class No. | Data No. |
|---|---|---|---|
| DSFall2019 | 2019/9 – 2020/1 | G5A/5C/5E | 913 |
| DSSpring2020 | 2020/3 – 2020/7 | G5A-5F | 518 |
| DSFall2020 | 2020/9 – 2021/1 | G6A-6F | 888 |
| Note: The number of DSSpring2020 is less than the other two datasets because of the COVID-19 pandemic which reduces the change of speaking English to the Kebbi Air. | | | |

Table I shows the valid collected data set from Rende elementary school from Sept. 9 to Jan. 2021. The students from three grade-5 classes (5A, 5C, and 5E) first joined this co-learning class from 2019/9 to 2020/1, and class 5B routinely joined in the next semester (2020/3 to 2020/7). The same students were advanced from G5 to G6 in Sept. 2020; hence all of the students of G6A-G6F joined this program from 2020/9 to 2021/1. Figs. 9(a)-9(c) show the accumulated *RAA*-recognized scores collected at classes from 2019/9 to 2020/1, 2020/3 to 2020/7, and 2020/9 to 2021/1, respectively. It indicates that their English speaking ability has improved because the growth in *RAA*-recognized scores is significant from Sept. 2019 to Jan. 2021.

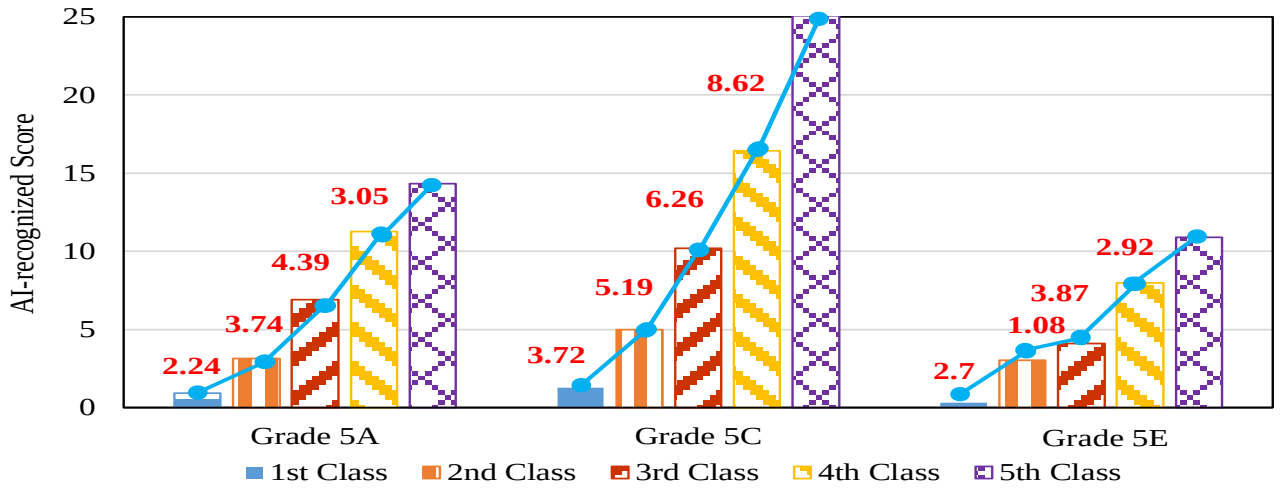

(a)

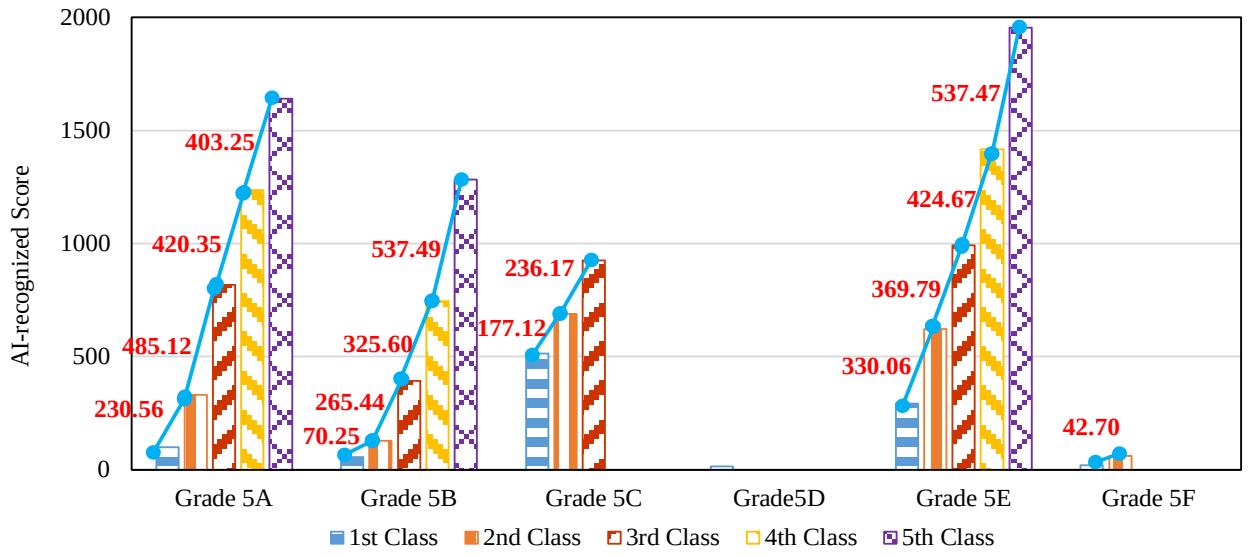

(b)

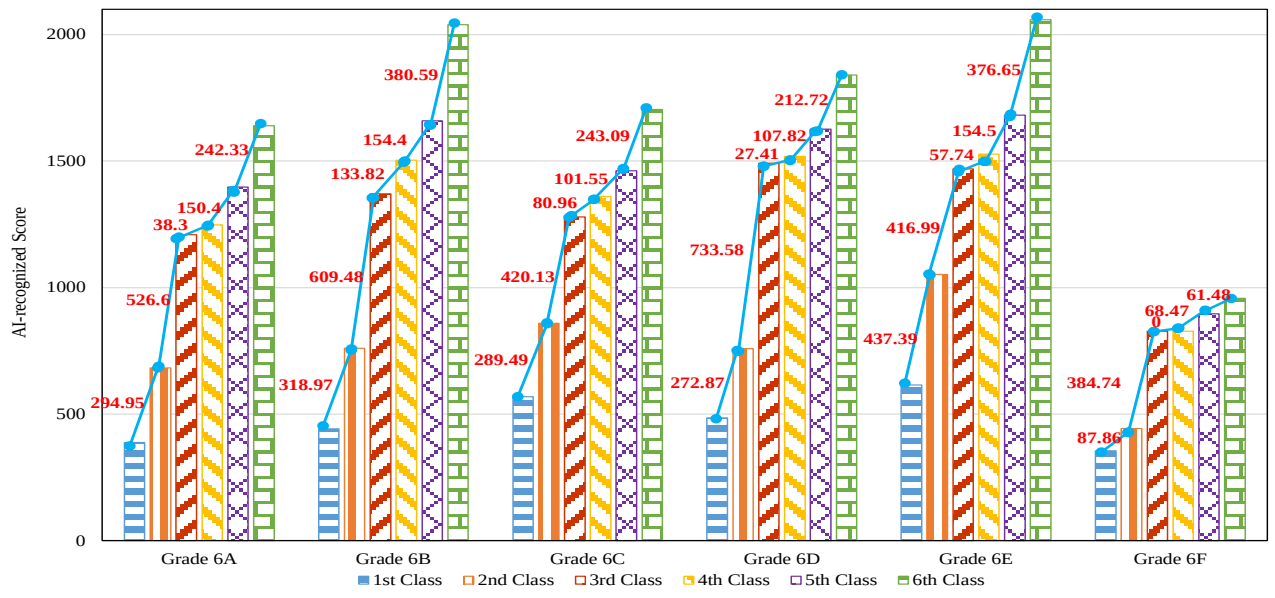

(c)

Fig. 9. Accumulated *RAA*-recognized scores during (a) 2019/9 – 2020/1, (b) 2020/3 – 2020/7, and (c) 2020/9 – 2021/1.

## B. *ISEA Data Analysis based on Deep Learning*

We used the data collected at English Co-Learning Class for the deep learning model construction. The adopted deep learning mechanism has nine feature values, $x_1$, $x_2$, …, and $x_9$, and a label value, $y$, for each student in each class [11], where 1) $x_1$ is school code, $x_2$ is the grade of the student, and $x_3$ is the gender of the student, 2) $x_4$, $x_5$, and $x_6$ are scores extracted from his/her feedback, 3) $x_7$ denotes his/her number of speaking practice in class, 4) $x_8$ is the correctly *RAA*-recognized ratio, 5) $x_9$ is the correctly and partially *RAA*-recognized score, and 6) $y$ is his/her monthly English paper-and-pencil test score at school.

TABLE II. TRAINING DATA NO AND TESTING DATA NO.

| Data Set | Training Data No. | Testing Data No. | Total |
|---|---|---|---|
| DSFall2019 | 787 | 338 | 1125 |
| DSSpring2020 | 302 | 130 | 432 |
| DSFall2020 | 599 | 258 | 857 |

Table II shows the number of the training data and testing data before data cleaning. We designed our experiments 1, 2, and 3 as follows: 1) Datasets of Exps. 1, 2, and 3 are *DSFall2019*, *DSSpring2020*, and *DSFall2020*, respectively. 2) Epochs are 100, 200, 300, 400, and 500. Table III shows partial scaled training data of Exp. 1-4. Table IV shows the loss values of training data (Tra. Data), validation data (Val. Data), and testing data (Tst. Data) of three main experiments, Exps.1 to Exps. 3. In this paper, we use the MSE function to evaluate the learning performance, and the best performance is Exp. 1-4, Exp. 2-3, and Exp. 3-4, respectively, for the three experiments. The best MSE values for learning Tra. Data are shown in a bold font format. Figs. 10(a)-10(c) show the loss curves in Exps. 1-4, 2-3, and 3-4, respectively.

TABLE III. PARTIAL SCALED TRAINING DATA OF EXP. 1-4.

| Input Features | | | | | | | | | Output |
|---|---|---|---|---|---|---|---|---|---|
| $x_1$ | $x_2$ | $x_3$ | $x_4$ | $x_5$ | $x_6$ | $x_7$ | $x_8$ | $x_9$ | $y$ |
| 1 | 5 | 1 | 0.848 | 0.792 | 0.833 | 0.25 | 0.226 | 0.23 | 0.819 |
| 1 | 5 | 0 | 0.708 | 0.82 | 0.934 | 0.198 | 0.658 | 0.309 | 0.931 |
| 1 | 5 | 1 | 0.837 | 0.896 | 0.862 | 0.131 | 0.254 | 0.182 | 0.472 |
| 1 | 5 | 1 | 0.833 | 0.709 | 0.875 | 0.013 | 0 | 0 | 0.694 |
| 1 | 5 | 1 | 0.882 | 0.804 | 0.853 | 0.531 | 0.735 | 0.997 | 0.556 |
| 1 | 5 | 1 | 0.882 | 0.912 | 0.982 | 0.25 | 0.226 | 0.23 | 0.375 |
| 1 | 5 | 0 | 0.845 | 0.812 | 0.923 | 0.302 | 0.924 | 0.582 | 0.611 |
| | | | | | ⋮ | | | | |
| 1 | 5 | 1 | 0.644 | 0.74 | 0.991 | 0.135 | 0.687 | 0.216 | 0.556 |
| 1 | 5 | 0 | 0.841 | 0.705 | 0.925 | 0.417 | 0.714 | 0.666 | 0.986 |
| 1 | 5 | 1 | 0.913 | 0.975 | 0.899 | 0.198 | 0.282 | 0.105 | 1 |
| 1 | 5 | 1 | 0.901 | 0.709 | 0.978 | 0.356 | 0.413 | 0.339 | 1 |

TABLE IV. LOSS VALUES OF THREE MAIN EXPERIMENTS.

| Data Set | Exp. No | Epochs | Mean Square Error (MSE) | | |
|---|---|---|---|---|---|
| | | | Tra. Data | Val. Data | Tst. Data |
| DSFall2019 | Exp. 1-1 | 100 | 0.059 | 0.066 | 0.064 |
| | Exp. 1-2 | 200 | 0.054 | 0.067 | 0.064 |
| | Exp. 1-3 | 300 | 0.052 | 0.067 | 0.068 |
| | Exp. 1-4 | 400 | **0.049** | 0.067 | 0.076 |
| | Exp. 1-5 | 500 | 0.057 | 0.073 | 0.071 |
| DSSpring2020 | Exp. 2-1 | 100 | 0.063 | 0.074 | 0.059 |
| | Exp. 2-2 | 200 | 0.039 | 0.053 | 0.041 |
| | Exp. 2-3 | 300 | **0.030** | 0.042 | 0.040 |
| | Exp. 2-4 | 400 | 0.031 | 0.045 | 0.041 |
| | Exp. 2-5 | 500 | 0.035 | 0.048 | 0.042 |
| DSFall2020 | Exp. 3-1 | 100 | 0.042 | 0.057 | 0.042 |
| | Exp. 3-2 | 200 | 0.041 | 0.055 | 0.043 |
| | Exp. 3-3 | 300 | 0.040 | 0.055 | 0.042 |
| | Exp. 3-4 | 400 | **0.037** | 0.055 | 0.042 |
| | Exp. 3-5 | 500 | 0.038 | 0.059 | 0.042 |

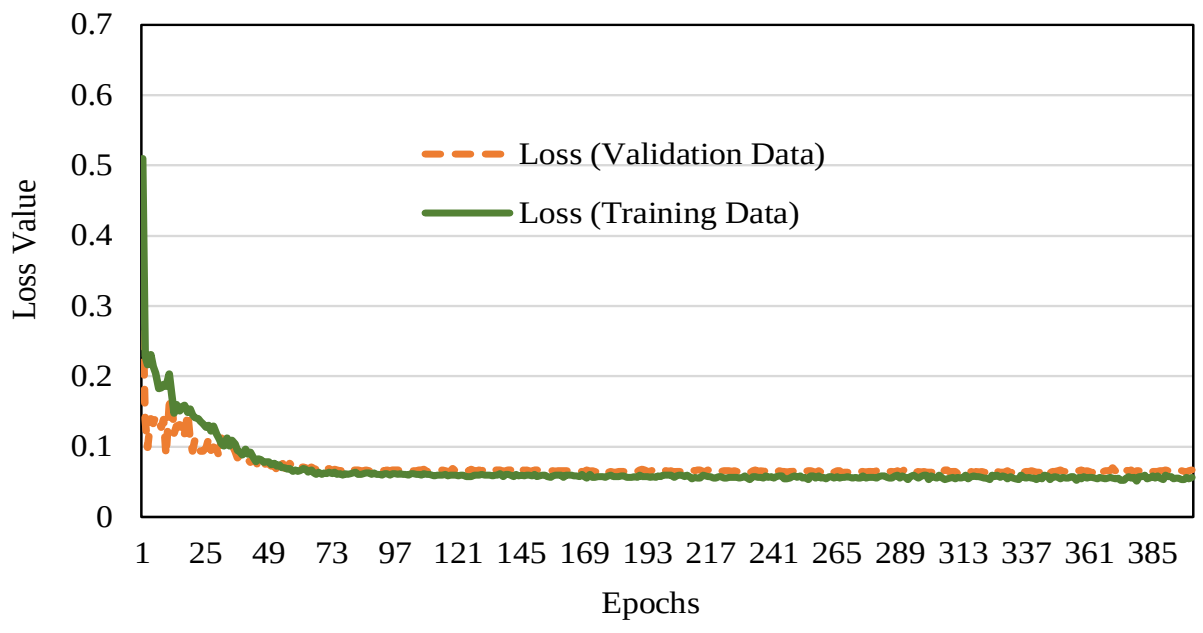

(a)

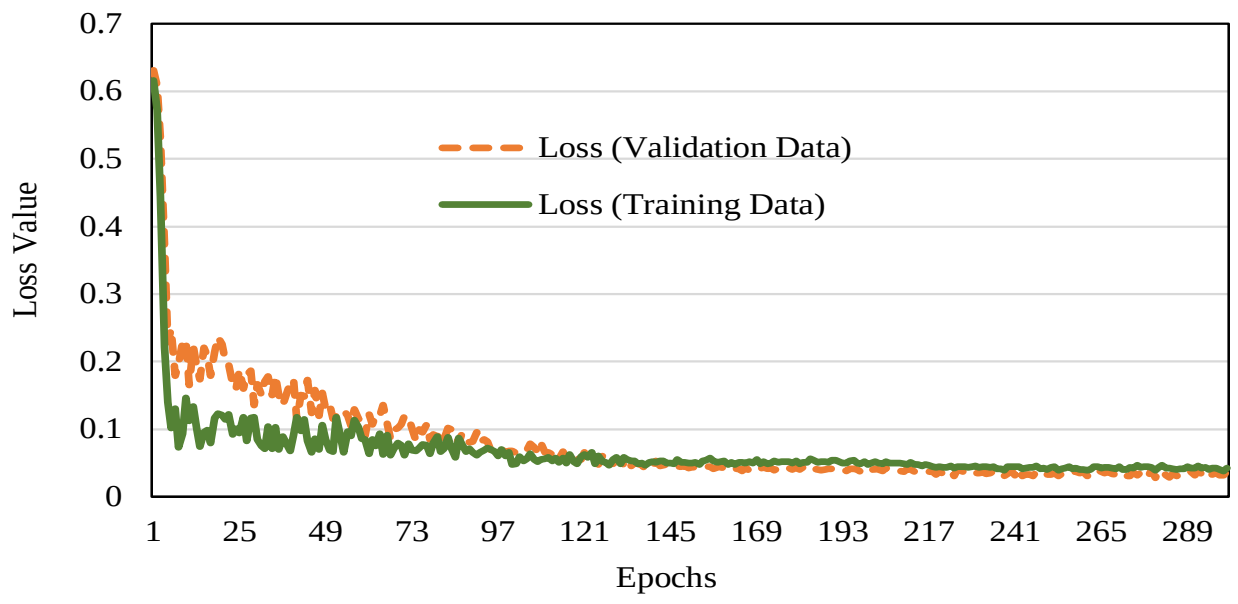

(b)

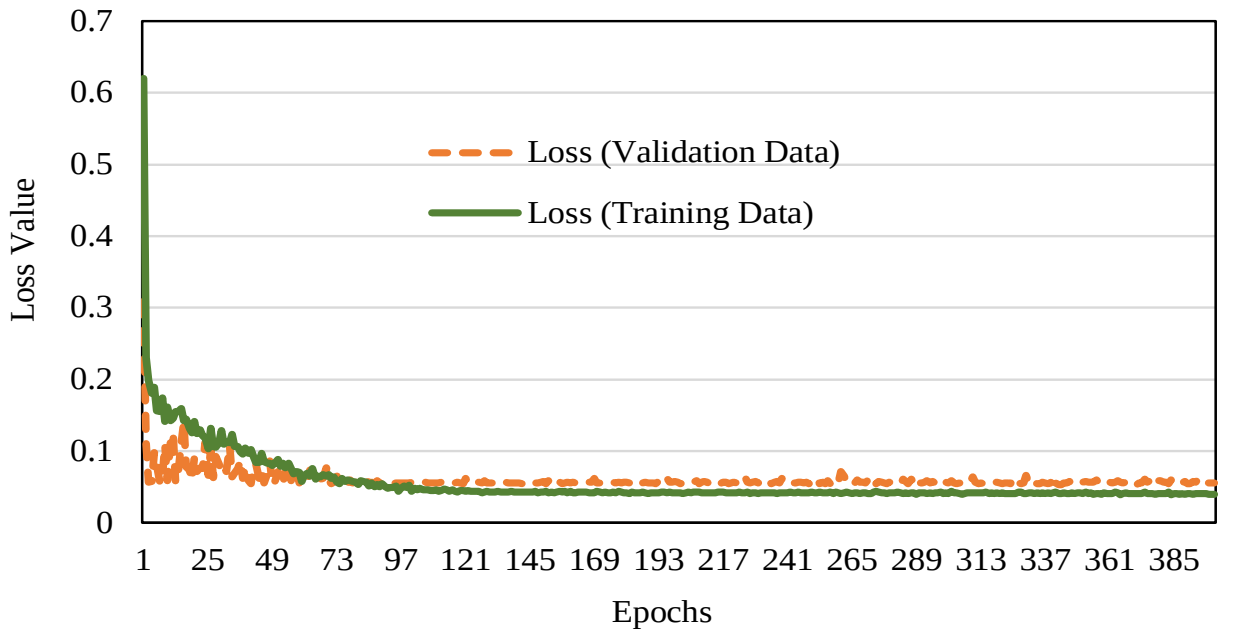

(c)

Fig. 10. Loss curves in (a) Exp. 1-4 with 400 epochs, (b) Exp. 2-3 with 300 epochs, and (c) Exp. 3-4 with 400 epochs.

## *C. Feedback from Elementary-School Students*

There are 281 students from 11 classes, including grade-5 five classes and grade-6 six classes, who joined this AI-FML course at Rende elementary school in Taiwan. To understand more about the involved student's learning state, we design a feedback form for the AI-FML course, shown in Fig. 11. Each student made a response to eight questions in the last class of Fall Semester 2020.

**Learning Feedback for AI-FML Human and Machine Co-Learning Course**

Name:_________ Grade:_________ Class:_________ No:_________

**Q1***: What did you learn from AI-FML Human and Machine Co-learning course?*
(Note: You can have multiple choices)
(1) *AI-FML Operation Platform:* Human Knowledge, Logic Operation Rule, Fuzzy Markup language, Machine Learning, Fuzzy Inference, or Fitness/Accuracy
(2) *AIoT:* Blockly program for Kebbi Air, Blockly program for Zenbo, or AI-FML Agent
(3) *Computer Operation:* PowerPoint, Excel, Film Editing, or Copy/Paste

**Q2:** *What does the robot need to predict flooding, the recommendation for travel, and preference for McDonald's?* (Note: You can have multiple choices) (Note: You can have multiple choices)
Human Knowledge, Logic Operation Rule, Blockly Program, or AI-FML Agent

**Q3:** *Which robot do you like?* (Note: You can have multiple choices)
Zenbo or Kebbi Air

**Q4:** *How do you your robots to make them talk to you, sing/dance for you, and as intelligent as you?*

**Q5:** *What is the greatest help that this course does for you?*
(For example, you can predict preference for McDonald's, or you can design a Blockly program for your robots after taking this course.)

**Q6:** *Who you want to thank most? Why?*

**Q7:** *What else this course can be improved?*

**Q8:** *Will you want to take this course next semester? Why?*

Fig. 11. Form of learning feedback for AI-FML course.

Figs. 12(a) and 12(b) show the pie charts of grade-5 and grade-6 students' responses to Q8, respectively. It indicates that most of the grade-5 students welcome the AI-FML course. But, some of the grade-6 students (about 20%) don't want to take the AI-FML course next semester. After observing students' responses to Q8, we found that about 52% and 24% of students of Grade 6C responded "No" and "I have no idea" to Q8, respectively. The reasons for "*responding No to Q8*" includes "hoping to have more free time to play games in class," "teaching contents are too difficult or too boarding," "I will move to the other school next semester," and so on. Fig. 13 shows the average responded score to Q1 of eleven involved classes. The total options to Q1 are 13 so the maximum of Q1 is 13. It indicates that 1) most of the students consider themselves learning something after taking the AI-FML course, and 2) in terms of grade-6 students, class-6C children consider themselves learning least from the course which makes sense because 52% don't want to take the AI-FML course next semester.

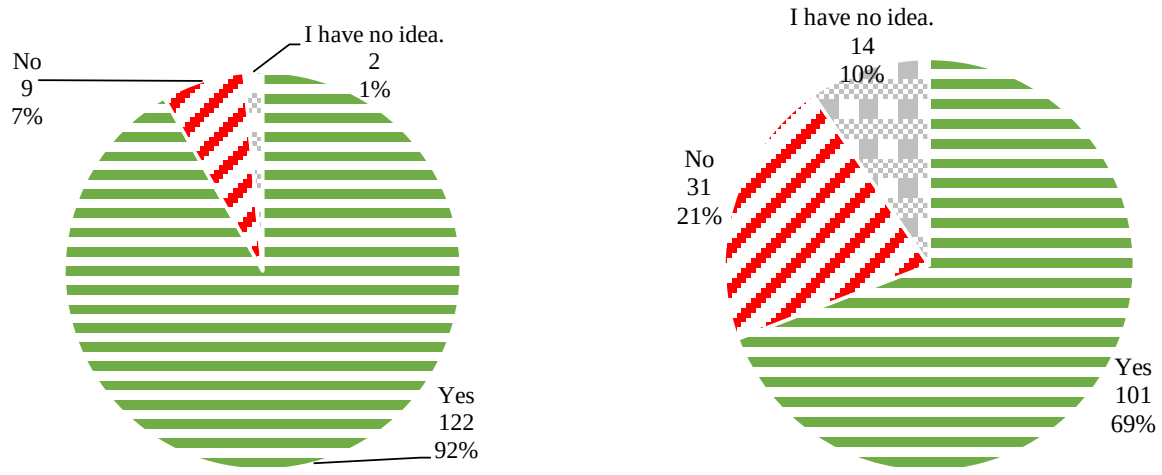

(a) (b)

Fig. 12. Pie charts of (a) grade-5 students' and (b) grade-6 students' response to Q8.

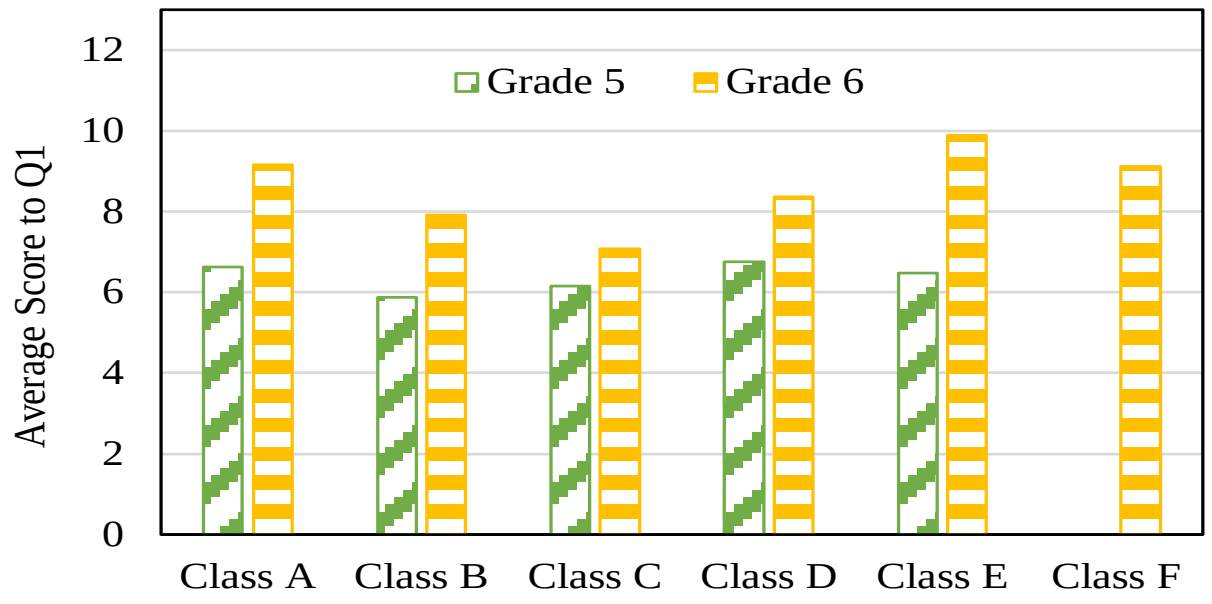

Fig. 13. Average responded score to Q1 of eleven involved classes

## *D. 2021 AI Winter Camp for Senior-High-School Students*

We applied this student and machine co-learning model to senior-high-school students. We cooperated with Taiwan AI Academy to hold an AI Winter Camp on the afternoon of Feb. 3, 2021. Table V shows the AI-FML learning and practice design for high school students at the 2021 AI winter camp in Taiwan. The total number of the students is 49 and they were grouped into six teams. In Part 3, the six-team students experienced in *AIoT-FML learning tools* like BCI music listening and *RAA* English speaking applications. The developed *RAA* hosted 50 people to experience and Table VI shows the statistics in average *RAA*-recognized score, count of correctly recognized by the *RAA*, and count of partially recognized by the *RAA*. It indicates that the average score for six teams is about 0.5 and it has a 1:1 ratio between the count of correctly and partially recognized by the *RAA*. Figs. 14(a) and 14(b) shows the 27-student feedback to "*Q7: Do you think the teaching contents is simple, common, or difficult?*" and "*Q8: Can you understand the* *teaching contents?*" Most students consider the teaching materials are common (not simple also not difficult) and can understand most of the teaching.

TABLE V. AI-FML LEARNING AND PRACTICE FOR 2021 AI WINTER CAMP.

| Time | Contents | |
|---|---|---|
| **Part 1: AI-FML Human and Machine Co-Learning** | | |
| 13:30-14:00 | **Section 1**: Demo for BCI Music Application<br>**Section 2**: Demo for *RAA* on English Speaking | |
| 14:00-14:50 | **Section 1**: Knowledge Base and Rule Base<br>**Section 2**: AI-FML Robot Connection<br>**Section 3**: Data Analysis and Machine Learning | |
| **Part 2: AI-FML Practice in Machine Learning** | | |
| 15:10-15:30 | BCI Music Application and *RAA* English Speaking | |
| **Part 3: Experience in *AIoT-FML learning tool* Application** | | |
| 15:30-15:50 | T1-T3 | T4-T6 |
| | BCI Music Listening | *RAA* English Speaking |
| 15:50-16:10 | *RAA* English Speaking | BCI Music Listening |
| 16:10-16:20 | Communication with robots Palro and Palmi | |
| **Part 4: Sharing Experience** | | |
| 16:20-16:30 | Learning experience and performance by video | |
| Note: For more details on the 2021 AI Winder camp for senior-high-school students, please connect to https://youtu.be/DkKMKTthLMM. | | |

TABLE VI. STATISTIC DATA FOR 2021 AI WINTER CAMP.

| Team No | T1 | T2 | T3 | T4 | T5 | T6 | Avg |
|---|---|---|---|---|---|---|---|
| Average *RAA*-recognized Score (Maximum is 1) | 0.512 | 0.556 | 0.595 | 0.641 | 0.297 | 0.513 | 0.519 |
| | Count | | | | | | |
| Correctly Recognized by *RAA* | 4 | 4 | 5 | 6 | 1 | 5 | 25 |
| Partially Recognized by *RAA* | 4 | 3 | 4 | 3 | 6 | 5 | 25 |

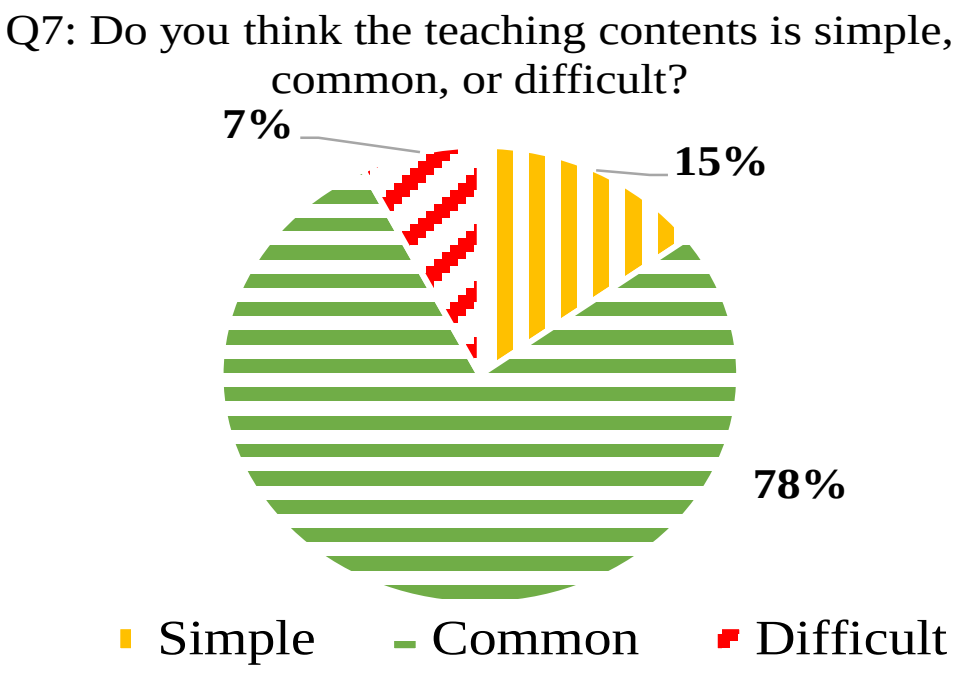

(a)

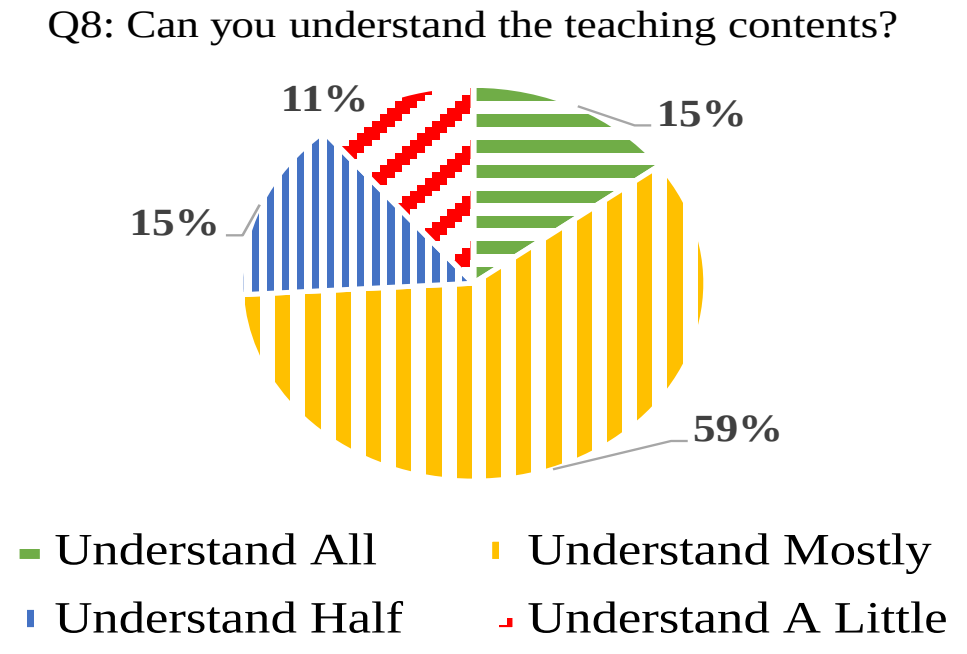

(b)

Fig. 14. Feedback to (a) Q7 and (b) Q8.

## V. CONCLUSIONS

This paper proposes a *Robotic Assistant Agent* (*RAA*) that can assist students and machines in co-learning English listening and AI-FML practice based on the *AIoT-FML learning tool*. The *RAA* is introduced to help students to learn human languages, such as English listening, and AI-FML language, such as Fuzzy Markup Language, to Rende elementary school in Taiwan from Sept. 2019 to Jan. 2021. Besides, we also present the novel AI-FML learning model with *AIoT-FML learning tool* at the 2021 AI Winter camp held in Taiwan AI Academy on Feb. 3. 2021. Moreover, the proposed *RAA* with AI-FML is first embedded in AIoT device to be a novel *AIoT-FML learning tool* in Taiwan and will promote to Japan and Europe for student future learning. From the experimental results and the collection of students' feedback, it shows that this kind of learning model for AI-FML future real-world applications and practice is suitable for elementary-school and high-school students, and the learning performance of elementary-school students is improved.

## ACKNOWLEDGMENT

The authors would like to thank the staff of the Center for Research of Knowledge Application & Web Service (KWS Center) of NUTN, members of OASE Lab. of NUTN, the involved faculty and students of Rende elementary school, Taiwan AI Academy in Taiwan, and IEEE CIS High School Outreach Subcommittee. Finally, we would like to thank the staff of Mediawave Intelligent Communication Limited. for their technical support for the GenioPy board.

## REFERENCE

[1] IEEE CIS, "What is CI?," May 2020. [Online] Available: https://cis.ieee.org/about/what-is-ci.

[2] C. T. Yang, "AIoT in Smart Application," *MOST Global Affairs and Science Engagement* (*GASE*)-*Taiwan Newsletter*, Aug. 2020.

[3] P. Zhang, R. P. Chen, T. Dai, Z. T. Wang, and K. Wu, "An AIoT-based system for real-time monitoring of tunnel construction," *Tunnelling and Underground Space Technology*, vol. 109, pp. 1-12, 2020.

[4] P. S. Chiu, J. W. Chang, M. C. Lee, C. H. Chen, and D. S. Lee, "Enabling intelligent environment by the design of emotionally aware virtual assistant: a case of smart campus," *IEEE Access*, vol. 8, pp. 62032-62041, 2020.

[5] C. S. Lee, Y. L. Tsai, M. H. Wang, W. K. Kuan, Z. H. Ciou, and N. Kubota, "AI-FML agent for robotic game of Go and AIoT real-world co-learning applications," *2020 World Congress on Computational Intelligence* (*IEEE WCCI 2020*), Glasgow, Scotland, UK, Jul. 19-24, 2020.

[6] Zsystem Technology Co., Genio Py STEM Board V2, Feb. 2021. [Online] Available: https://aiot.zsystem.com.tw/.

[7] C. S. Lee, M. H. Wang, Y. L. Tsai, W. S. Chang, M. Reformat, G. Acampora, and N. Kubota, "FML-based reinforcement learning agent with fuzzy ontology for human-robot cooperative edutainment," *International Journal of Uncertainty, Fuzziness and Knowledge-Based Systems*, vol. 28, no. 6, pp. 1023-1060, 2020.

[8] C. S. Lee, M. H. Wang, Y. L. Tsai, L. W. Ko, B. Y. Tsai, P. H. Hung, L. A. Lin, and N. Kubota, "Intelligent agent for real-world applications on robotic edutainment and humanized co-learning," *Journal of Ambient Intelligence and Humanized Computing*, vol. 11, pp. 3121-3139, 2019.

[9] FUZZ-IEEE 2021, "FML-based machine learning competition for human and smart machine co-learning on real-world applications @ FUZZ-IEEE 2021," Feb. 2021. [Online] Available: https://attend.ieee.org/fuzzieee-2021/competition/.

[10] C. S. Lee, Y. Nojima, G. Acampora, M. Reformat, N. Kubota, R. Saga, I. C. Wu, J. M. S. Hidalgo, "FML-based machine learning competition for human and smart machine co-learning on real-world applications @ FUZZ-IEEE 2021," Feb. 2021. [Online] Available: http://oase.nutn.edu.tw/fuzz2021-fmlcompetition/.

[11] C. S. Lee, M. H. Wang, W. K. Kuan, Z. H. Ciou, Y. L. Tsai, W. S. Chang, L. C. Li, N. Kubota, T. X. Huang, E. Sato-Shimokawara, and T. Yamaguchi, "A study on AI-FML robotic agent for student learning

behavior ontology construction," *2020 International Symposium on Community-centric Systems* (*CcS 2020*), Tokyo, Japan, Sept. 23-26, 2020.

[12] G. Acampora, "Fuzzy markup language: A XML based language for enabling full interoperability in fuzzy systems design," in G. Acampora, V. Loia, C. S. Lee, and M. H. Wang (editors), On the Power of Fuzzy Markup Language, Springer-Verlag, Germany, Jan. 2013, pp. 17-31.

[13] J. M. Soto-Hidalgo, Jose M. Alonso, G. Acampora, and J. Alcala-Fdez, "JFML: A Java library to design fuzzy logic systems according to the IEEE Std 1855-2016," IEEE Access, vol. 6, pp. 54952-54964, 2018.

[14] J. M. Soto-Hidalgo, A. Vitiello, J. M. Alonso, G. Acampora, J. Alcala-Fdez, "Design of fuzzy controllers for embedded systems with JFML," International Journal of Computational Intelligence Systems, vol. 12, no. 1, pp. 204-214, 2019.

[15] J. M. Soto-Hidalgo, J. M. Alonso, and J. Alcalá-Fdez, "Java Fuzzy Markup Language," Jan. 2019. [Oneline] Available: http://www.uco.es/JFML.